\documentclass[sigconf,nonacm]{acmart}
\AtBeginDocument{%
  }
\makeatletter
\def\@mkauthors@iii{%
  \hsize=\textwidth
  \global\setbox\mktitle@bx=\vbox{\noindent
    \unvbox\mktitle@bx\par\medskip\centering
    \lineskip=0.5pc\relax
    \@authorfont\arxivauthors\par
    \medskip
    \normalsize\normalfont\arxivaffiliations\par
    \smallskip
    \normalsize\normalfont\href{mailto:simonfan@mail.ustc.edu.cn}{simonfan@mail.ustc.edu.cn}\par
    \bigskip}}
\makeatother
\graphicspath{{figures/}}
\usepackage{enumitem}
\usepackage{multirow}
\usepackage{makecell}
\usepackage{colortbl}
\usepackage{placeins}
\usepackage{tabularx}
\usepackage[most]{tcolorbox}
\definecolor{baselineSingle}{HTML}{F4EFE4}
\definecolor{baselineTask}{HTML}{EAF1E8}
\definecolor{baselineConfidence}{HTML}{F0EBF5}
\definecolor{methodBlue}{HTML}{E5EFFA}
\DeclareRobustCommand{\baselinekey}[2]{{\setlength{\fboxsep}{1.2pt}\colorbox{#1}{\strut #2}}}
\newcommand{\arxivauthors}{%
Chenxiao Fan$^{1,2}$, Chongming Gao$^{1}$, Gangyi Zhang$^{2}$,
Yaxin Gong$^{1}$, Leyang Shen$^{3}$, Jiakai Wang$^{2}$,
\\Dong Wang$^{2}$, Yang Liu$^{2}$, Haoyan Liu$^{1}$,
Xinting Hu$^{1}$, Fuli Feng$^{1}$, Xiangnan He$^{1}$}
\newcommand{\arxivaffiliations}{%
$^{1}$University of Science and Technology of China\quad
$^{2}$Qwen Business Unit of Alibaba\quad
$^{3}$National University of Singapore}
\begin{document}
\title{Learning Calibrated Confidence for Multi-Agent Collaboration}
\author{Chenxiao Fan, Chongming Gao, Gangyi Zhang, Yaxin Gong, Leyang Shen, Jiakai Wang, Dong Wang, Yang Liu, Haoyan Liu, Xinting Hu, Fuli Feng, Xiangnan He}
\renewcommand{\shortauthors}{Chenxiao Fan et al.}
\begin{abstract}

Complex information seeking on the Web often requires collecting evidence from multiple documents and reasoning across them, as in multi-hop question answering.
Multi-agent systems address such tasks by dividing work among specialized agents, where upstream agents produce intermediate answers and supporting evidence that downstream agents build upon.
However, upstream answers vary in correctness, and most systems provide no quantitative signal about it, leaving the downstream agent to judge whether to trust or verify what it receives based solely on the textual evidence.

In this paper, we propose MASCAL. Each agent learns to produce a correctness probability for its answer by jointly optimizing task performance and confidence calibration in reinforcement learning; the upstream agent passes this probability at handoff, giving the downstream agent a quantitative basis for its decisions, and the final answer also carries a calibrated probability, quantifying the reliability of the system output.
To guide the downstream agent in using this signal during training, we introduce confidence-aware weighting, which increases the learning weight at positions responsive to the upstream probability, so that the agent learns more from decisions where it actively uses the signal.
Across three collaboration architectures spanning Web retrieval and reasoning tasks, MASCAL improves task accuracy and calibration quality for both agents, and generalizes to deeper collaboration chains and different model scales.
Intervention analysis confirms that upstream confidence affects downstream decision behavior, and the learned final confidence supports selective answering and targeted rechecking.

\end{abstract}

\maketitle
\section{Introduction}

Complex information seeking on the Web often requires collecting evidence from multiple sources and reasoning across them, as in multi-hop question answering~\cite{hotpotqa,musique} or cross-document fact verification~\cite{hover}.
Large language model (LLM) agents have made substantial progress on such tasks by combining Web retrieval with step-by-step reasoning~\cite{searchr1,toolwild}.
However, a single agent remains limited in both coverage and reliability when the task spans many evidence sources or requires extended reasoning chains~\cite{anchorrag,graphcogent}.
Multi-agent systems (MAS) address this through role specialization: different agents retrieve separate pieces of evidence, answer different sub-questions, or produce candidate answers for another role to review~\cite{mindsearch,opera,malt,colacare}.
In a typical pipeline, an upstream agent produces an intermediate answer with supporting evidence and hands off a report to a downstream agent for further processing~\cite{pamas}.
The quality of the final output therefore depends both on each role's subtask performance and on how the downstream agent handles the upstream conclusions it receives~\cite{gong2026upstream}.

\begin{figure}[t]
\centering
\includegraphics[width=\columnwidth]{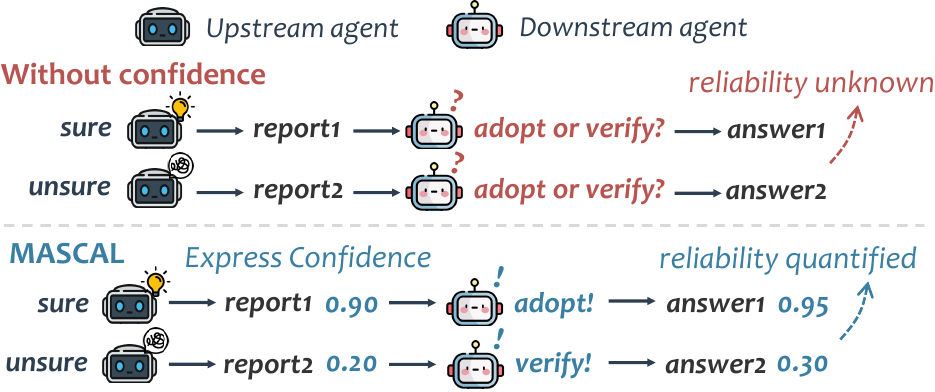}
\caption{Confidence at the agent handoff. Without a confidence signal, the downstream agent cannot distinguish a reliable report from an unreliable one. MASCAL attaches calibrated confidence to each report and to the final answer, guiding downstream decisions and enabling reliability assessment of the system output.}
\label{fig:intro_handoff}
\end{figure}

Upstream agents vary in quality across inputs, and the downstream agent must assess each report it receives.
It must decide whether to trust the upstream conclusion and move forward, or to launch additional searches to verify~\cite{tosearch}.
Yet as Figure~\ref{fig:intro_handoff} illustrates, most current MAS convey only the answer and textual evidence, giving the downstream agent no quantitative signal about answer correctness to inform this choice.
If the upstream agent could convey, alongside its answer, a probability reflecting how likely that answer is to be correct, the downstream agent would gain a concrete basis for its decision.
This raises the question: can each agent learn to express \emph{calibrated answer confidence} while learning its task, and can the downstream agent learn to act on this signal?

Training with calibration objectives has shown that a single LLM can learn to produce well-calibrated correctness probabilities for its answers, providing a basis for introducing reliable confidence signals into multi-agent settings~\cite{rlcr,dcpo,credo}.
In multi-agent settings, existing work has explored confidence expression and utilization in debates, discussions, and pairwise interactions~\cite{lacie,dmad,talkmesh}, while other studies have investigated cross-role uncertainty propagation and system-level probability estimation~\cite{propuq,cage}.
These efforts each address part of the problem, but no existing framework connects confidence learning, transmission, and utilization across the full collaboration pipeline: the upstream agent needs to learn calibrated confidence for its own answer, and the downstream agent needs to learn to incorporate this signal into its decisions.

To bridge this gap, we propose \textbf{MASCAL}, in which each agent outputs a correctness probability for its answer alongside completing its subtask; the upstream agent passes this probability to the downstream agent together with its intermediate conclusions.
This probability is learned by incorporating a calibration objective into reinforcement learning: task performance and confidence calibration are jointly optimized within a single training loop, so that the model produces probabilities matching its actual correctness rates.
To further guide the downstream agent in leveraging this signal, we introduce \emph{confidence-aware weighting}, which measures how much each generation position responds to the upstream probability and increases the learning weight of the most responsive positions, so that the agent learns more from decisions where it actively uses the upstream signal.
The downstream agent also outputs a calibrated probability for its final answer, allowing the system to withhold uncertain outputs or direct rechecking where it is most needed.

We evaluate MASCAL under three collaboration architectures (sequential handoff, report fusion, and answer review) spanning Web-based multi-hop question answering, logical reasoning, and mathematical reasoning.
MASCAL improves task accuracy over single-agent baselines, MAS without confidence transmission, and existing confidence-based approaches, while the confidence signals of both upstream and downstream agents show improved calibration quality.
Intervention analysis further shows that upstream confidence has a measurable effect on downstream decisions, and the learned final confidence supports coverage-accuracy control through selective answering and targeted rechecking.
In generalization experiments, models trained on a two-hop architecture transfer directly to deeper Web-retrieval chains, and the same recipe remains effective at smaller model scales.

Our contributions are as follows:
\begin{itemize}[leftmargin=*]
\item To enable calibrated confidence communication in multi-agent collaboration, we propose \textbf{MASCAL}, which integrates calibration objectives into reinforcement learning so that each agent learns to produce a calibrated correctness probability for its answer, and further introduces \emph{confidence-aware weighting} to guide the downstream agent's use of upstream confidence.
\item Across three collaboration architectures spanning Web retrieval and reasoning tasks, we demonstrate that calibrated confidence transmission improves task accuracy and calibration quality in MAS, and that upstream confidence measurably affects downstream decisions. MASCAL further generalizes to deeper collaboration chains and different model scales.
\item We show that the learned final confidence supports selective answering and targeted rechecking, reducing the error rate of delivered outputs substantially, or directing additional computation to where it is most needed.
\end{itemize}

\section{Related Work}

In this section, we review related work along two lines: how agents in MAS learn to collaborate and how confidence and uncertainty signals have been introduced into multi-agent settings.

\subsection{Multi-Agent Collaboration and Learning}

LLM-driven MAS have been widely applied to Web tasks requiring multi-step reasoning and evidence gathering~\cite{mindsearch,opera,graphragr1}.
At the orchestration level, one line of work learns to select collaboration modes and role configurations per query~\cite{maas,masrouter,daao,ofamas}.
Given a fixed role assignment, another line focuses on training each role's policy, curating data from successful and failed trajectories~\cite{malt,sirius} or designing rewards that combine correctness with interaction behavior~\cite{maporl}.
Scaling up to the system level, other methods jointly improve multi-role performance~\cite{magrpo,mmoarag,ma4div} through terminal reward propagation~\cite{mhgpo}, preference objectives~\cite{sysdpo,optima}, and inter-role evaluation signals~\cite{comas}.
In Web retrieval settings, confidence has been used to control the expansion of query decomposition trees, balancing answer quality against search cost~\cite{prunerag}.

These methods improve MAS task performance through various mechanisms, yet their training objectives center on final-answer correctness or system efficiency. The reliability of handoff content, specifically whether the upstream agent's answer is likely correct, remains outside the scope of learning. MASCAL incorporates calibrated answer confidence into role-level training, making reliability judgments part of the information agents exchange.

\begin{figure*}[t]
\centering
\includegraphics[width=\textwidth]{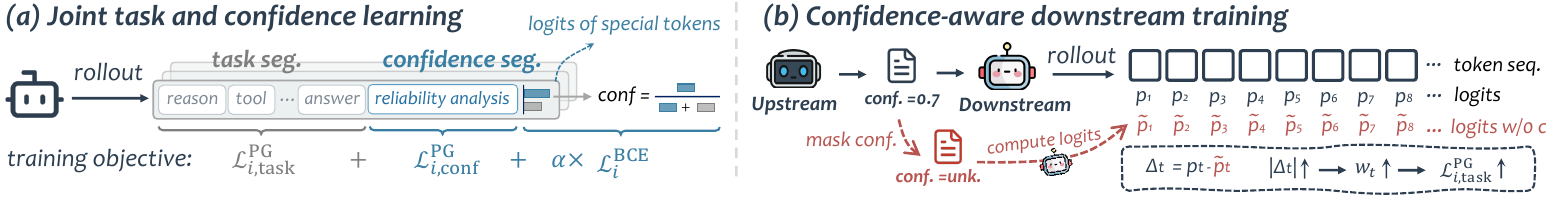}
\caption{The MASCAL training pipeline. (a)~Each rollout is split into a task segment and a confidence segment, trained with separate policy gradients and a BCE loss on the readout $c$. (b)~Masking the upstream confidence reveals which downstream tokens depend on it; those positions are up-weighted during task learning.}
\label{fig:method}
\end{figure*}

\subsection{Confidence and Uncertainty in MAS}

Calibrated reinforcement learning has recently shown that a single LLM can learn to produce well-calibrated correctness probabilities for its answers, providing a foundation for introducing reliable confidence signals into MAS~\cite{rlcr,dcpo,ugr,credo}.
In multi-agent settings, early attempts used self-reported confidence for weighted voting~\cite{reconcile} or injected uncertainty into peer prompts to modulate answer updates in debate~\cite{debunc}.
These untrained signals, however, have been found unreliable~\cite{uala}, sometimes causing unstable interaction behavior.
Subsequent work has therefore begun to train confidence expression and utilization: incorporating the listener's behavior into the speaker's confidence learning~\cite{lacie}, using correctness labels to train verbalized confidence and update behavior in debate~\cite{dmad,udpo}, or learning confidence scorers for speaker selection and listener filtering~\cite{talkmesh}. These methods focus primarily on debate and discussion, where all participants address the same question.

When collaboration goes beyond same-question debate and involves multi-step reasoning or role-level handoffs, the scope of uncertainty broadens. Along this direction, work has studied how to propagate uncertainty through reasoning steps~\cite{saup}, let downstream agents inherit uncertainty from predecessor decisions~\cite{propuq}, or use uncertainty to trigger task re-delegation in hierarchical systems~\cite{hassum}. At the output end, counterfactual calibration over communication graphs~\cite{cage} and multi-source confidence composition~\cite{composition} have been used to improve final-answer probability quality.

The methods above introduce confidence into MAS from complementary angles: expression training, signal utilization, cross-role propagation, and final-answer probability estimation. Each, however, addresses only part of the pipeline, and no existing framework connects confidence production, transmission, and utilization within a unified learning process. MASCAL fills this gap: each role learns to produce calibrated confidence through calibration RL, the downstream agent trains on reports actually generated by the upstream to incorporate this signal into its decisions, and confidence-aware weighting further strengthens this utilization.

\section{Preliminary}

In this section, we formalize the multi-agent collaboration setting and the notion of calibrated answer confidence, and introduce the policy optimization framework used throughout our method.

\subsection{Task Formulation}

We consider collaborative systems where a complex task $x$ is decomposed into subtasks handled by multiple LLM agents in sequence.
Each agent $i$ receives its subtask $t_i$, initial context $o_i$ (e.g., retrieved documents), and a set of upstream reports $\mathcal{M}_i$, produces a trajectory $\tau_i$ containing reasoning actions, an answer $a_i$, and supporting rationale $e_i$, and passes a report to the downstream agent:
\begin{equation}
\label{eq:report}
m_i = (t_i,\; a_i,\; e_i).
\end{equation}
The final agent in the chain produces the system's answer to the original task $x$, based on its own subtask and all upstream reports.

To enable each agent to communicate a judgment of answer correctness at handoff, we augment the report with answer confidence $c_i \in [0,1]$, the agent's estimate of the probability that $a_i$ is correct:
\begin{equation}
\label{eq:report_conf}
m_i = (t_i,\; a_i,\; e_i,\; c_i).
\end{equation}
We aim for $c_i$ to be calibrated with respect to actual correctness~\cite{guo2017calibration}:
\begin{equation}
\label{eq:calibration}
P\bigl(V(t_i, a_i) = 1 \;\big|\; c_i = p\bigr) \approx p, \quad \forall\, p \in [0, 1],
\end{equation}
where $V(t_i, a_i) \in \{0, 1\}$ is a verification function that determines the correctness of $a_i$.
Calibration is especially important in this multi-agent setting: unlike a single agent reporting to a human user, $c_i$ is consumed automatically by another agent and directly affects its decisions. A miscalibrated probability can systematically bias downstream behavior, making calibration a prerequisite for effective confidence communication.

\subsection{Group Relative Policy Optimization}

We adopt Group Relative Policy Optimization (GRPO)~\cite{deepseekmath} as our training framework. GRPO does not require a separate critic network; instead, it estimates advantages directly from rewards within each sampling group. For each input, the policy $\pi_\theta$ samples $G$ trajectories, and the advantage of each is the standardized deviation of its reward from the group mean:
\begin{equation}
\label{eq:grpo_advantage}
\hat{A}_{gi} = \frac{r_{gi} - \bar{r}_g}{\mathrm{std}(r_g) + \epsilon},
\end{equation}
where $r_{gi}$ is the reward for the $i$-th trajectory in group $g$, and $\bar{r}_g$, $\mathrm{std}(r_g)$ are the group mean and standard deviation. The policy is updated via an importance-weighted objective:
\begin{equation}
\label{eq:grpo_loss}
\mathcal{L}_{\mathrm{GRPO}} = -\frac{1}{G}\sum_{i=1}^{G} \frac{1}{|\tau_i|}\sum_{t=1}^{|\tau_i|} \frac{\pi_\theta(z_t \mid z_{<t}, x)}{\pi_{\theta_{\mathrm{old}}}(z_t \mid z_{<t}, x)} \, \hat{A}_{gi},
\end{equation}
where $\pi_{\theta_{\mathrm{old}}}$ is the policy used to sample the trajectories. Because the advantage computation depends only on the reward function's output, different reward signals can each produce their own advantages from the same group of rollouts, which provides the basis for the multi-signal training described next.

\section{Method: MASCAL}
\label{sec:method}

In this section, we present MASCAL (Figure~\ref{fig:method}). We first describe how each agent learns to output calibrated confidence alongside its task answer, and then introduce confidence-aware downstream training, which trains the downstream agent to leverage the upstream confidence signal in its decisions.

\subsection{Joint Task and Confidence Learning}
\label{sec:joint_learning}

In a standard MAS, each agent generates a trajectory containing reasoning actions and an answer. MASCAL additionally requires each agent to produce a natural-language confidence analysis assessing the reliability of its own answer.
The full trajectory of agent $i$ is split into a \emph{task segment} $\tau_i^{\mathrm{task}}$ (reasoning, tool calls, and the answer) and a \emph{confidence segment} $\tau_i^{\mathrm{conf}}$ (reliability analysis; see Figure~\ref{fig:method}a):
\begin{equation}
\label{eq:trajectory}
\tau_i = [\tau_i^{\mathrm{task}};\; \tau_i^{\mathrm{conf}}].
\end{equation}
This segmentation carries over to training: task and confidence policy gradients are computed over their respective segments, keeping calibration learning separate from task behavior~\cite{dcpo,credo}.

At the end of the confidence segment, the confidence value is read out from the logits of two designated tokens H and L~\cite{credo,selfref}:
\begin{equation}
\label{eq:readout}
c_i = \sigma(z_\mathrm{H} - z_\mathrm{L}),
\end{equation}
where $\sigma$ is the sigmoid function. Unlike verbalized confidence, which requires sampling the probability as discrete tokens, this differentiable readout allows gradients from a probability loss to propagate directly, providing finer-grained calibration supervision than reward signals alone.

Each agent is trained with three learning signals: the task policy gradient drives reasoning behavior, the confidence policy gradient shapes the quality of the analysis text, and a direct probability loss calibrates the numerical readout. We describe each below.

\paragraph{Task policy gradient.}
The task reward guides reasoning and tool use. Let $Y_i = V(t_i, a_i)$ denote correctness and $F$ format compliance:
\begin{equation}
\label{eq:task_reward}
r^{\mathrm{task}}_i = F \cdot Y_i.
\end{equation}
The corresponding GRPO advantage (Eq.~\ref{eq:grpo_advantage}) is applied only to the task segment $\tau_i^{\mathrm{task}}$ of each trajectory.

\paragraph{Confidence policy gradient.}
The confidence reward measures how close the readout is to correctness when the output format is valid:
\begin{equation}
\label{eq:conf_reward}
r^{\mathrm{conf}}_i = F \cdot \bigl(1 - (Y_i - c_i)^2\bigr).
\end{equation}
Its policy gradient is applied only to $\tau_i^{\mathrm{conf}}$, encouraging the model to produce analyses that better distinguish correct from incorrect answers, which in turn makes the logit readout more reliable.

\paragraph{Direct probability supervision.}
A BCE loss aligns the confidence readout directly with the correctness label:
\begin{equation}
\label{eq:bce}
\mathcal{L}^{\mathrm{BCE}}_{i} = -\frac{1}{|\mathcal{V}_i|}\sum_{j \in \mathcal{V}_i} \bigl[ Y_{ij} \log c_{ij} + (1 - Y_{ij}) \log (1 - c_{ij}) \bigr],
\end{equation}
where $\mathcal{V}_i$ is the set of trajectories with valid confidence readouts. Gradients propagate directly through the readout, calibrating the probability value independently of the generated text.

The confidence policy gradient and the BCE loss complement each other: the former trains the model to write analyses that better separate correct from incorrect answers, and the latter directly adjusts the probability value through the readout. The joint objective for agent $i$ combines all three:
\begin{equation}
\label{eq:role_objective}
\mathcal{L}_i = \mathcal{L}^{\mathrm{PG}}_{i,\mathrm{task}} + \mathcal{L}^{\mathrm{PG}}_{i,\mathrm{conf}} + \alpha \, \mathcal{L}^{\mathrm{BCE}}_{i},
\end{equation}
where $\alpha$ controls the weight of direct probability supervision.

\subsection{Confidence-Aware Downstream Training}
\label{sec:downstream_training}

With the upstream agent trained to express calibrated confidence, the downstream agent needs to learn in an environment where it actually consumes these signals. The downstream agent is trained with the same joint objective (Eq.~\ref{eq:role_objective}):
\begin{equation}
\label{eq:downstream_obj}
\min_{\theta_B} \; \mathbb{E}_{x \sim D_B, \, \mathcal{M} \sim \Pi_A(\cdot \mid x)} \bigl[ \mathcal{L}_B(\theta_B;\, x,\, \mathcal{M}) \bigr],
\end{equation}
where $D_B$ is the downstream training set and $\Pi_A$ the report distribution of the frozen upstream.

Standard policy gradients apply the same task advantage to every position and cannot distinguish which decisions are influenced by the upstream confidence. If a token's generation probability changes substantially when the upstream confidence is present versus absent, the model is relying on the confidence signal at that position. Strengthening the learning signal at these positions rewards the pattern of incorporating confidence into correct decisions, turning an implicit dependency into an explicit training objective. We operationalize this through \emph{confidence-aware weighting}.

Given a downstream trajectory, we replace the upstream confidence in the report with an unavailability marker and re-score the trajectory with the same model (Figure~\ref{fig:method}b). The resulting per-position log-probability difference is:
\begin{equation}
\label{eq:delta}
\Delta_t = \log P_\theta(z_t \mid x, z_{<t}) - \log P_\theta(z_t \mid x_{\backslash c},\, z_{<t}),
\end{equation}
where $x_{\backslash c}$ is the input with confidence masked and $z_{<t}$ the generated tokens before position $t$. The absolute difference is normalized and centered around the trajectory mean to produce weights:
\begin{equation}
\label{eq:weight}
s_t = \tanh(|\Delta_t|), \qquad w_t = 1 + \lambda \, (s_t - \bar{s}),
\end{equation}
with $\bar{s}$ the trajectory-level mean and $\lambda$ controlling the adjustment magnitude. The weighted task advantage is:
\begin{equation}
\label{eq:weighted_advantage}
\widetilde{A}^{\mathrm{task}}_{git} = \mathrm{stopgrad}(w_{git}) \cdot \hat{A}^{\mathrm{task}}_{gi},
\end{equation}
$\widetilde{A}^{\mathrm{task}}_{git}$ replaces the original task advantage in $\mathcal{L}^{\mathrm{PG}}_{B,\mathrm{task}}$ (Eq.~\ref{eq:role_objective}), so that positions more responsive to upstream confidence receive correspondingly stronger task learning signals. The weighting applies only to the task policy gradient; the confidence policy gradient and BCE loss are both left unchanged.

For the downstream agent to encounter realistic confidence signals during training, including cases where the upstream is uncertain or incorrect, it must train on reports produced by the actual upstream policy. The upstream agent is therefore first trained with the joint objective (Eq.~\ref{eq:role_objective}) and then frozen to generate confidence-bearing reports. The downstream agent trains on these reports with the same objective, with its task policy gradient additionally modulated by confidence-aware weighting, ensuring that the report distribution seen during training matches deployment.

\begin{figure}[t]
\centering
\includegraphics[width=0.9\columnwidth]{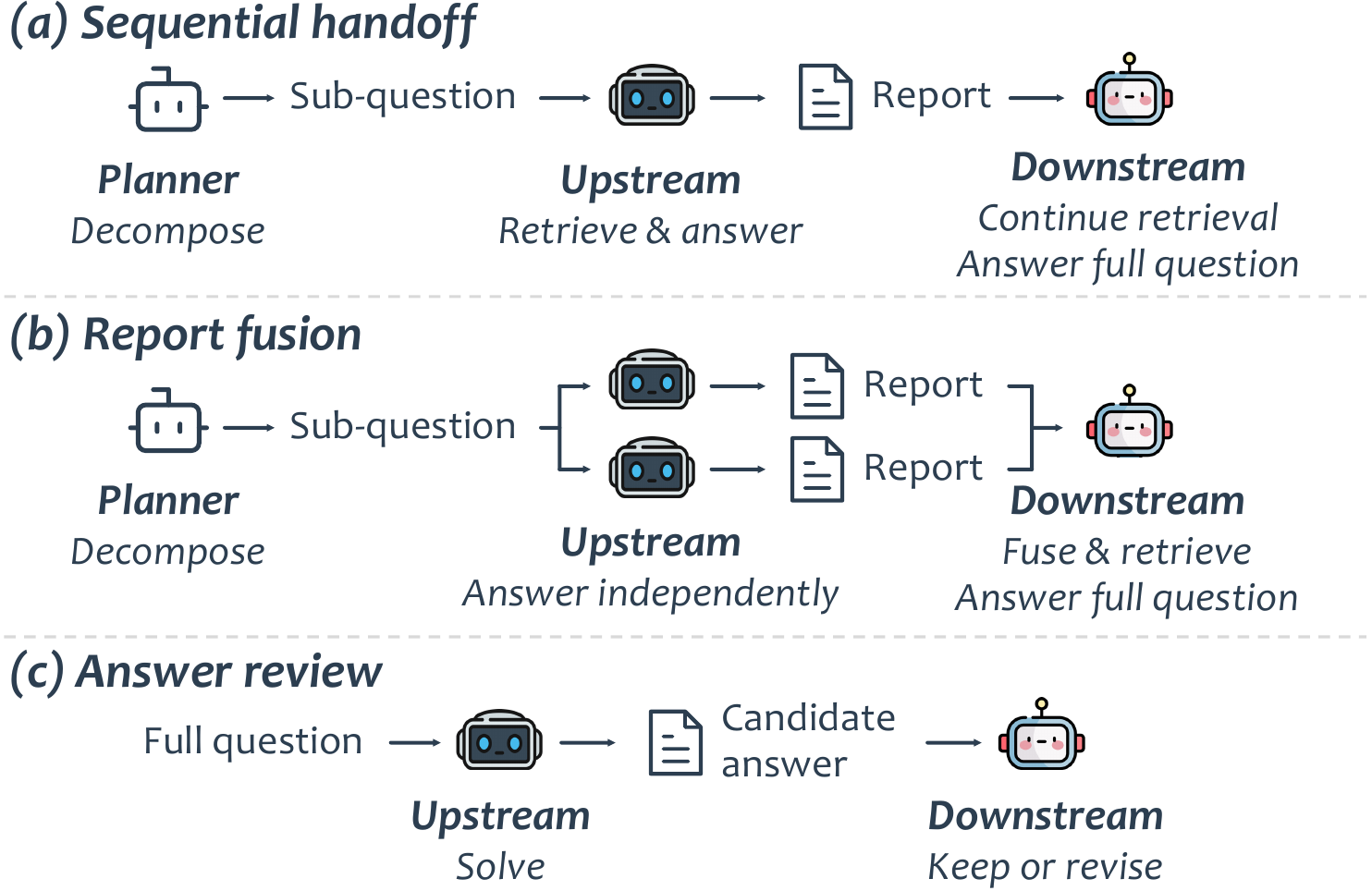}
\caption{Collaboration architectures in our experiments.}
\label{fig:collaboration_architectures}
\end{figure}

\begin{table}[t]
\centering
\caption{Dataset splits and question counts.}
\label{tab:datasets}
\footnotesize
\setlength{\tabcolsep}{2pt}
\renewcommand{\arraystretch}{1.03}
\begin{tabular*}{\columnwidth}{@{\extracolsep{\fill}}llc@{\hspace{8pt}}lr@{}}
\toprule
Task & Main dataset & Train / Val. / Test & \multicolumn{2}{c}{Additional tests ($n$)} \\
\midrule
\multirow{2}{*}{\makecell[l]{Multi-hop\\QA}}
 & \multirow{2}{*}{MuSiQue}
 & \multirow{2}{*}{12,938 / 719 / 1,252}
 & HotpotQA & 1,000 \\
 & & & 2Wiki & 1,000 \\
\midrule
\multirow{2}{*}{Logic}
 & \multirow{2}{*}{LogiQA2}
 & \multirow{2}{*}{9,000 / 500 / 1,572}
 & ReClor & 500 \\
 & & & AR-LSAT & 230 \\
\midrule
\multirow{2}{*}{Math}
 & \multirow{2}{*}{DeepScaleR}
 & \multirow{2}{*}{9,000 / 500 / 500}
 & MATH500 & 500 \\
 & & & AMC \& AIME & 181 \\
\bottomrule
\end{tabular*}
\end{table}

\begin{table*}[t]
\centering
\caption{Task accuracy (\%) comparing MASCAL with \baselinekey{baselineSingle}{single-agent}, \baselinekey{baselineTask}{MAS without confidence}, and \baselinekey{baselineConfidence}{confidence-based MAS} baselines. Best results are \textbf{bolded}.}
\label{tab:main}
\setlength{\tabcolsep}{3pt}
\renewcommand{\arraystretch}{1.18}
\small
\resizebox{\textwidth}{!}{%
\begin{tabular}{l *{16}{c}}
\toprule
& \multicolumn{4}{c}{\textbf{Sequential handoff}} & \multicolumn{4}{c}{\textbf{Report fusion}} & \multicolumn{4}{c}{\textbf{Answer review (logic)}} & \multicolumn{4}{c}{\textbf{Answer review (math)}} \\
\cmidrule(lr){2-5} \cmidrule(lr){6-9} \cmidrule(lr){10-13} \cmidrule(lr){14-17}
Method & MuSiQue & HotpotQA & 2Wiki & Avg. & MuSiQue & HotpotQA & 2Wiki & Avg. & LogiQA2 & ReClor & AR-LSAT & Avg. & MATH500 & AMC \& AIME & DeepScaleR & Avg. \\
\midrule
\rowcolor{baselineSingle}
Base Single-Agent & 24.65 & 42.67 & 40.90 & 36.07 & 24.65 & 42.67 & 40.90 & 36.07 & 69.94 & 80.30 & 35.65 & 61.96 & 83.20 & 37.43 & 60.60 & 60.41 \\
\rowcolor{baselineSingle}
Task-RL Single-Agent & 47.26 & 52.27 & 51.87 & 50.47 & 47.26 & 52.27 & 51.87 & 50.47 & 76.91 & 85.65 & 52.07 & 71.54 & 91.30 & 51.31 & 75.70 & 72.77 \\
\addlinespace[2pt]
\rowcolor{baselineTask}
Base MAS & 30.35 & 43.93 & 50.27 & 41.52 & 27.69 & 46.77 & 51.37 & 41.94 & 68.73 & 78.10 & 34.84 & 60.56 & 82.85 & 37.50 & 60.35 & 60.23 \\
\rowcolor{baselineTask}
Task-RL MAS & 43.32 & 51.90 & 47.27 & 47.49 & 45.07 & 58.27 & 61.00 & 54.78 & 77.53 & 86.25 & 57.77 & 73.85 & 91.30 & 51.38 & 76.05 & 72.91 \\
\rowcolor{baselineTask}
MAGRPO & 44.12 & 53.43 & 49.07 & 48.87 & 43.93 & 57.77 & 67.20 & 56.30 & 73.58 & 82.95 & 49.73 & 68.75 & 88.80 & 48.96 & 73.40 & 70.39 \\
\rowcolor{baselineTask}
MHGPO & 36.61 & 50.83 & 60.43 & 49.29 & 36.29 & 52.27 & 63.13 & 50.56 & 71.36 & 80.75 & 42.77 & 64.96 & 90.35 & 50.55 & 74.65 & 71.85 \\
\addlinespace[2pt]
\rowcolor{baselineConfidence}
ConfMAD & 34.58 & 47.93 & 57.83 & 46.78 & 34.58 & 51.13 & 59.33 & 48.35 & 71.33 & 81.30 & 37.50 & 63.38 & 77.45 & 27.76 & 51.90 & 52.37 \\
\rowcolor{baselineConfidence}
DMAD & 45.50 & 55.87 & 48.00 & 49.79 & 40.36 & 51.87 & 58.03 & 50.09 & 74.25 & 83.10 & 51.14 & 69.50 & 86.40 & 42.06 & 67.45 & 65.30 \\
\addlinespace[2pt]
\rowcolor{methodBlue}
\textbf{MASCAL} & \textbf{49.31} & \textbf{60.57} & \textbf{67.13} & \textbf{59.00} & \textbf{47.98} & \textbf{62.23} & \textbf{74.13} & \textbf{61.45} & \textbf{78.05} & \textbf{88.25} & \textbf{63.48} & \textbf{76.59} & \textbf{92.90} & \textbf{62.36} & \textbf{80.80} & \textbf{78.69} \\
\bottomrule
\end{tabular}%
}
\end{table*}

\section{Experiments}
\label{sec:experiments}

In this section, we first describe the experimental setup, then organize our analysis around five research questions:
\begin{itemize}[leftmargin=*]
\item \textbf{RQ1.} Does calibrated confidence transmission improve task performance in multi-agent collaboration?
\item \textbf{RQ2.} Is upstream confidence well calibrated, and how does it influence downstream decisions?
\item \textbf{RQ3.} Is final-answer confidence well calibrated, and can it support selective answering and targeted rechecking?
\item \textbf{RQ4.} What does each design component contribute?
\item \textbf{RQ5.} Does MASCAL generalize to deeper collaboration chains and different model scales?
\end{itemize}

\subsection{Experimental Setup}
\label{sec:setup}

\subsubsection{Collaboration architectures.}
We evaluate MASCAL and the multi-agent baselines under the three architectures in Figure~\ref{fig:collaboration_architectures}:
\begin{itemize}[leftmargin=*,nosep]
\item \emph{Sequential handoff.} A planner decomposes the original question. The upstream agent retrieves evidence, answers the sub-question, and hands off its report. The downstream agent continues retrieving and answers the original question.
\item \emph{Report fusion.} After the planner decomposes the question, two upstream agents independently answer the same sub-question and each delivers a report; the downstream agent synthesizes both and proceeds.
\item \emph{Answer review.} The upstream agent answers the full question and delivers a candidate answer; the downstream agent reviews and either retains or revises it.
\end{itemize}

\subsubsection{Datasets and metrics.}
Our primary evaluation uses multi-hop QA~\cite{musique,hotpotqa,twowiki}, where agents retrieve evidence from Web documents and reason across them to answer complex questions. These questions naturally decompose into sub-questions, making them suitable for sequential handoff and report fusion, where each agent handles a separate retrieval step.
We further evaluate on logical reasoning with reading comprehension questions~\cite{logiqa2,reclor,arlsat}, and mathematical reasoning with competition-level problems~\cite{math}. These tasks involve a single reasoning process that is difficult to decompose, so we evaluate them under answer review, where one agent produces a candidate solution and the other examines and revises it.
For each task type, we train on one dataset and evaluate on both its in-domain test set and additional out-of-domain datasets to assess generalization. Table~\ref{tab:datasets} summarizes the splits.

We report task accuracy (Acc.) and assess confidence quality using the Brier score~\cite{proper_scoring} and AUROC~\cite{fawcett2006roc}:
\begin{equation}
\begin{aligned}
\mathrm{Brier} &= \frac{1}{N}\sum_{i=1}^{N}(c_i-y_i)^2,\\
\mathrm{AUROC} &= \Pr(c^+>c^-)+\tfrac12\Pr(c^+=c^-),
\end{aligned}
\end{equation}
where $c_i$ is the agent's confidence for the $i$-th answer with binary correctness label $y_i$, and $c^+$, $c^-$ denote the confidence on correct and incorrect answers, respectively.

\subsubsection{Models and baselines.}
All agents are initialized from Qwen3-8B~\cite{qwen3}. We compare MASCAL against three groups of baselines:
(1) \emph{Single-agent}: the base model and its RL-trained variant, each solving the task without collaboration.
(2) \emph{MAS without confidence}: agents collaborate under the same architectures but transmit no confidence signal, including an untrained and an RL-trained MAS, as well as MAGRPO~\cite{magrpo} and MHGPO~\cite{mhgpo}, which optimize multi-role performance through cooperative MARL and terminal reward propagation, respectively.
(3) \emph{Confidence-based MAS}: ConfMAD derives confidence from sequence-level probabilities with post-hoc calibrators~\cite{confmad}; DMAD trains verbalized confidence expression and its use through correctness rewards~\cite{dmad}.

\subsubsection{Implementation.}
We split the training data equally into disjoint subsets: one trains the upstream agent, which is then frozen to generate reports for downstream training on the other.
For downstream training, we use a learning rate of $2\times10^{-6}$, $G=8$ rollouts per input, three epochs, and confidence-aware weighting strength $\lambda=0.5$. The calibration loss weight is $\alpha=0.03$ on MuSiQue and $0.01$ on LogiQA2 and DeepScaleR.
Retrieval experiments use a total search budget of three calls.
Baselines were tuned from published settings, with validation performance for hyperparameter selection.

Full experimental settings are provided in Appendix~\ref{sec:experimental_details}.

\subsection{Main Results (RQ1)}
\label{sec:main_results}

Table~\ref{tab:main} compares MASCAL with single-agent, multi-agent, and confidence-based baselines across four evaluation scenarios. MASCAL achieves the highest accuracy in all cases. The gains over Task-RL MAS, which shares the same architecture but transmits no confidence, are most pronounced in search-based collaboration, where an explicit reliability signal gives the downstream agent richer information for guiding retrieval and reasoning.

In sequential handoff, Task-RL MAS falls below the single agent: the downstream agent inherits one upstream report with no indication of its reliability, and uninformed collaboration hurts rather than helps. By attaching calibrated confidence at the handoff, MASCAL raises accuracy well above the single-agent level.

Other multi-agent approaches do not match MASCAL's improvements. Cooperative RL methods (MAGRPO, MHGPO) jointly optimize multiple roles but establish no reliability communication between them. MAGRPO outperforms Task-RL MAS on average under both retrieval architectures, but neither method improves consistently across the four scenarios. Confidence-based methods (ConfMAD, DMAD) do transmit a reliability signal, but post-hoc calibration and verbalized expressions yield limited accuracy gains over Task-RL MAS; on answer review, both fall below it. These results indicate that the confidence signal should be learned through calibration training alongside the task, and the downstream agent should also learn to act on it when making decisions.

\FloatBarrier
\subsection{Confidence Quality and Its Effect on Downstream Agents (RQ2)}
\label{sec:rq2}

This section analyzes the calibration quality of the upstream confidence signal and how it affects downstream behavior.

\begin{table}[!ht]
\centering
\captionsetup{skip=4pt}
\caption{Upstream confidence quality. Scores are averaged over test sets for logic and math (Brier$\downarrow$, AUROC$\uparrow$).}
\label{tab:upstream_confidence}
\footnotesize
\setlength{\tabcolsep}{2pt}
\renewcommand{\arraystretch}{1.12}
\begin{tabularx}{\columnwidth}{@{}l*{6}{>{\centering\arraybackslash}X}@{}}
\toprule
& \multicolumn{2}{c}{MuSiQue} & \multicolumn{2}{c}{Logic} & \multicolumn{2}{c}{Math} \\
\cmidrule(lr){2-3}\cmidrule(lr){4-5}\cmidrule(l){6-7}
Method & Brier & AUROC & Brier & AUROC & Brier & AUROC \\
\midrule
Base MAS & 0.1496 & 0.7666 & 0.3472 & 0.6316 & 0.3726 & 0.7973 \\
Task-RL MAS & 0.1576 & 0.7887 & 0.1952 & 0.7081 & 0.2513 & 0.8361 \\
MASCAL-Verbal\footnotemark & 0.1440 & 0.6527 & 0.2403 & 0.5013 & 0.1092 & 0.7482 \\
\rowcolor{methodBlue}
\textbf{MASCAL} & \textbf{0.0972} & \textbf{0.8663} & \textbf{0.1516} & \textbf{0.7500} & \textbf{0.0827} & \textbf{0.8951} \\
\bottomrule
\end{tabularx}
\end{table}
\footnotetext{MASCAL-Verbal replaces the logit-based readout with verbalized confidence in the generated text, as in calibration RL methods such as RLCR and DCPO~\cite{rlcr,dcpo}.}

\textbf{Upstream confidence quality.}
Table~\ref{tab:upstream_confidence} evaluates the upstream agent's confidence on its own subtask answers. The base model produces poorly calibrated confidence, with Brier scores exceeding 0.34 on logic and math.
Task-only RL training improves AUROC, suggesting that task optimization can implicitly produce some discriminative signal, but Brier scores remain well above MASCAL's, so the probability values are not reliable.
MASCAL-Verbal shows unstable calibration across tasks, performing near MASCAL on some but dropping to chance-level AUROC on logic. Verbalized confidence goes through discrete token sampling, which introduces variance at inference and blocks direct gradient-based calibration during training.
MASCAL achieves the best Brier and AUROC across all settings, indicating that combining the logit-based readout with calibration training is essential for reliable upstream confidence.

\begin{table}[!ht]
\centering
\captionsetup{skip=4pt}
\caption{Accuracy (\%) and Brier score under upstream confidence interventions on a trained MASCAL system.}
\label{tab:confidence_interventions}
\footnotesize
\setlength{\tabcolsep}{3pt}
\renewcommand{\arraystretch}{1.10}
\begin{tabularx}{\columnwidth}{@{}l*{4}{>{\centering\arraybackslash}X}@{}}
\toprule
& \multicolumn{2}{c}{Multi-hop QA} & \multicolumn{2}{c}{Answer review} \\
\cmidrule(lr){2-3}\cmidrule(l){4-5}
Confidence & Handoff & Fusion & Logic & Math \\
\midrule
\rowcolor{methodBlue}
Original & \textbf{59.00} / \textbf{0.1947} & \textbf{61.45} / \textbf{0.1940} & \textbf{76.59} / \textbf{0.1489} & \textbf{78.69} / \textbf{0.0816} \\
Hidden & 50.00 / 0.3041 & 54.98 / 0.3251 & 76.28 / 0.2237 & 77.42 / 0.1646 \\
Random & 49.32 / 0.3055 & 54.80 / 0.3580 & 73.86 / 0.2470 & 73.07 / 0.2621 \\
\bottomrule
\end{tabularx}
\end{table}

\textbf{Confidence interventions.}
To isolate the causal effect of upstream confidence, we replace it with an unavailability marker (Hidden) or uniform random values (Random) at test time while keeping all other inputs unchanged (Table~\ref{tab:confidence_interventions}).
In search-based collaboration, both interventions cause substantial accuracy drops and Brier degradation, confirming that the downstream agent relies on the confidence signal when making retrieval and reasoning decisions.
Answer review shows a different pattern: hiding confidence leads to only minor accuracy drops, but replacing it with random values causes a clear degradation (math accuracy falls from 78.69\% to 73.07\%). When the signal is marked unavailable, the downstream agent falls back to content-based judgment and maintains accuracy. Random values, however, actively mislead its decisions, confirming that the model has learned to incorporate upstream confidence into its reasoning. Both interventions also degrade the downstream agent's own Brier scores across all scenarios, showing that upstream confidence shapes not only task accuracy but also the calibration of the downstream agent's output.

\begin{figure}[!ht]
\centering
\includegraphics[width=\columnwidth]{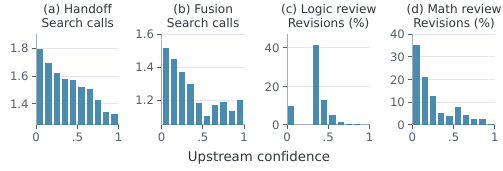}
\caption{MASCAL's behavior across ten confidence bins: additional search calls and answer revision rates. Fusion uses mean upstream confidence.}
\label{fig:confidence_behavior}
\end{figure}
\textbf{Downstream behavior.}
Figure~\ref{fig:confidence_behavior} bins MASCAL's downstream actions by upstream confidence. In search scenarios, lower confidence leads to more additional retrieval calls; in answer review, lower confidence is associated with higher revision rates. Together with the intervention results, these patterns confirm that calibrated confidence systematically shapes downstream decisions.

\FloatBarrier
\subsection{Final Confidence Quality and Utility (RQ3)}
\label{sec:rq3}

This section evaluates the confidence quality of the final answer and its practical utility for selective answering and targeted rechecking.

\begin{table}[!ht]
\centering
\captionsetup{skip=4pt,justification=raggedright,singlelinecheck=false}
\caption{Final confidence quality for \baselinekey{baselineSingle}{single-agent}, \baselinekey{baselineTask}{MAS without confidence}, \baselinekey{baselineConfidence}{confidence-based MAS}, and \baselinekey{methodBlue}{MASCAL variants}, averaged over test sets (Brier$\downarrow$, AUROC$\uparrow$).}
\label{tab:final_confidence}
\scriptsize
\setlength{\tabcolsep}{2pt}
\renewcommand{\arraystretch}{1.18}
\begin{tabularx}{\columnwidth}{@{}l*{4}{>{\centering\arraybackslash}X}@{}}
\toprule
& \multicolumn{2}{c}{Multi-hop QA} & \multicolumn{2}{c}{Answer review} \\
\cmidrule(lr){2-3}\cmidrule(l){4-5}
Method & Handoff & Fusion & Logic & Math \\
\midrule
\rowcolor{baselineSingle}
Base Single-Agent & \multicolumn{2}{c}{0.3689 / 0.7262} & 0.3472 / 0.6316 & 0.3726 / 0.7973 \\
\rowcolor{baselineSingle}
Task-RL Single-Agent & \multicolumn{2}{c}{0.4275 / 0.6873} & 0.1952 / 0.7081 & 0.2513 / 0.8361 \\
\addlinespace[2pt]
\rowcolor{baselineTask}
Base MAS & 0.4111 / 0.6528 & 0.3999 / 0.6671 & 0.3008 / 0.7183 & 0.3560 / 0.7655 \\
\rowcolor{baselineTask}
Task-RL MAS & 0.4498 / 0.6272 & 0.3965 / 0.5920 & 0.2266 / 0.6929 & 0.2608 / 0.8106 \\
\addlinespace[2pt]
\rowcolor{baselineConfidence}
ConfMAD & 0.3933 / 0.6957 & 0.3936 / 0.6937 & 0.3339 / 0.5616 & 0.4645 / 0.5778 \\
\rowcolor{baselineConfidence}
DMAD & 0.2919 / 0.7839 & 0.2859 / 0.7351 & 0.3023 / 0.5196 & 0.3440 / 0.5255 \\
\addlinespace[2pt]
\rowcolor{methodBlue}
MASCAL-Verbal & 0.3022 / 0.6965 & 0.3200 / 0.6857 & 0.2496 / 0.5192 & 0.1170 / 0.7600 \\
\rowcolor{methodBlue}
MASCAL-Unweighted\footnotemark & 0.2212 / 0.7297 & 0.1970 / 0.7408 & 0.1517 / 0.7816 & 0.0865 / \textbf{0.8937} \\
\rowcolor{methodBlue}
\textbf{MASCAL} & \textbf{0.1947} / \textbf{0.8183} & \textbf{0.1940} / \textbf{0.7701} & \textbf{0.1489} / \textbf{0.7870} & \textbf{0.0816} / 0.8930 \\
\bottomrule
\end{tabularx}
\end{table}

\textbf{Final-answer confidence.}
Table~\ref{tab:final_confidence} reports Brier scores and AUROC for the final answer's confidence across all methods.
MASCAL achieves the best or near-best scores in every scenario.
Among MAS without confidence methods, Task-RL MAS produces higher Brier scores on search tasks than the single agent (0.4498 vs. 0.4275 on handoff). Collaboration introduces additional uncertainty, and without a calibration mechanism, the downstream agent's confidence quality degrades rather than improves.
Both confidence-based baselines retain moderate discrimination on search tasks, but their AUROC drops to 0.52--0.58 on review, where neither post-hoc calibration (ConfMAD) nor debate-trained verbalized confidence (DMAD) produces signals that generalize to the handoff collaboration setting.
MASCAL-Verbal similarly shows unstable calibration quality across tasks, consistent with the upstream analysis in RQ2.
MASCAL-Unweighted also achieves strong calibration, but MASCAL further improves Brier and AUROC in most scenarios through its confidence-aware weighting.
\footnotetext{MASCAL-Unweighted removes only the confidence-aware weighting (Section~\ref{sec:downstream_training}) from downstream training.}

\begin{figure}[!ht]
\centering
\includegraphics[width=\columnwidth]{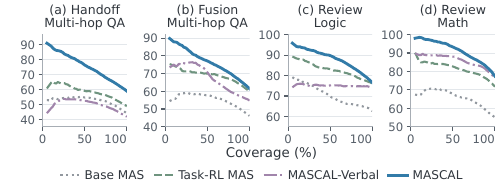}
\caption{Selective answering by final confidence. Accuracy is measured on the most confident answers at each coverage level, then averaged over test sets.}
\label{fig:selective_answering}
\end{figure}

\textbf{Selective answering.}
If final confidence is well calibrated, a system can improve output quality by answering only the questions it is most confident about~\cite{selectivenet}. Figure~\ref{fig:selective_answering} ranks answers by confidence and plots accuracy at each coverage level.
MASCAL's curve lies above all other methods at every coverage level across all four scenarios. At 20\% coverage, accuracy reaches 86--97\% depending on the task, well above full-coverage levels, confirming that the confidence signal effectively concentrates correct answers in the high-confidence range.
The other methods show substantially smaller accuracy spreads between low and full coverage, and MASCAL-Verbal is nearly flat on logic, indicating that its verbalized confidence carries almost no useful answer-ranking information for selective answering.
These results demonstrate that calibrated final confidence gives the system the ability to actively control output quality: in scenarios requiring high reliability, reducing coverage substantially lowers the overall error rate.

\begin{table}[!ht]
\centering
\captionsetup{skip=4pt}
\caption{Targeted rechecking in MASCAL under a fixed budget: $\Delta$Acc. (pp) when the budget is spent on the least-confident $k$\% of answers versus a random $k$\%.}
\label{tab:targeted_rechecking}
\footnotesize
\setlength{\tabcolsep}{3pt}
\renewcommand{\arraystretch}{1.12}
\begin{tabularx}{\columnwidth}{@{}l*{4}{>{\centering\arraybackslash}X}@{}}
\toprule
& Initial & \multicolumn{3}{c}{Rechecking budget} \\
\cmidrule(l){3-5}
Scenario & Acc. (\%) & 5\% & 10\% & 20\% \\
\midrule
Handoff & 59.00 & +0.15 / -0.04 & +0.37 / -0.09 & +0.74 / -0.17 \\
Fusion & 61.45 & +0.24 / +0.02 & +0.39 / +0.05 & +0.66 / +0.09 \\
\bottomrule
\end{tabularx}
\end{table}

\textbf{Targeted rechecking.}
Selective answering improves reliability by withholding uncertain answers, but real deployments usually need to return an answer for every query. Beyond deciding which answers to drop, confidence can also indicate where to invest additional computation. Our search-based collaboration already operates under a fixed per-question search budget, and a natural question is whether this budget can be routed to the answers most likely to be wrong. We study this on the handoff and fusion architectures by sorting answers by final confidence, sending the least-confident fraction back through the same MAS pipeline for one more round of retrieval, and comparing the accuracy gain against a random fraction of equal size (Table~\ref{tab:targeted_rechecking}).
Low-confidence rechecking improves accuracy monotonically with the budget on both architectures. Random rechecking stays near zero and even turns negative on handoff, since additional retrieval on already-confident answers can overwrite correct ones. The gap widens with the budget, indicating that final confidence directs the extra search to the cases that most need it.
Together with selective answering, this shows that calibrated final confidence is actionable: it decides not only which answers to return, but also where additional search should be spent.

\FloatBarrier
\subsection{Ablation Study (RQ4)}
\label{sec:ablation}

We isolate the contribution of each training component by removing it individually from the downstream agent's training (Table~\ref{tab:training_ablation}).

\begin{table}[!ht]
\centering
\captionsetup{skip=4pt}
\caption{Training ablations across four scenarios. Each pair reports Acc. (\%) / Brier$\downarrow$, averaged over test sets.}
\label{tab:training_ablation}
\scriptsize
\setlength{\tabcolsep}{2pt}
\renewcommand{\arraystretch}{1.18}
\begin{tabularx}{\columnwidth}{@{}l*{4}{>{\centering\arraybackslash}X}@{}}
\toprule
& \multicolumn{2}{c}{Multi-hop QA} & \multicolumn{2}{c}{Answer review} \\
\cmidrule(lr){2-3}\cmidrule(l){4-5}
Variant & Handoff & Fusion & Logic & Math \\
\midrule
\rowcolor{methodBlue}
\textbf{MASCAL} & \textbf{59.00} / \textbf{0.1947} & \textbf{61.45} / \textbf{0.1940} & \textbf{76.59} / \textbf{0.1489} & \textbf{78.69} / \textbf{0.0816} \\
\quad w/o weighting & 55.45 / 0.2212 & 57.26 / 0.1970 & 73.92 / 0.1517 & 78.32 / 0.0865 \\
\quad w/o calibration loss & 52.37 / 0.3092 & 51.81 / 0.4419 & 74.47 / 0.2413 & 78.05 / 0.0929 \\
\quad w/o confidence reward & 57.55 / 0.2102 & 54.46 / 0.2306 & 74.01 / 0.1538 & 78.08 / 0.0888 \\
MASCAL-Verbal & 49.97 / 0.3022 & 56.30 / 0.3200 & 73.86 / 0.2496 & 77.42 / 0.1170 \\
Task-RL MAS & 47.49 / 0.4498 & 54.78 / 0.3965 & 73.85 / 0.2266 & 72.91 / 0.2608 \\
\bottomrule
\end{tabularx}
\end{table}

The calibration loss has the largest effect on confidence quality, with its removal causing the most severe Brier degradation across scenarios. On fusion, removing calibration loss drops accuracy below Task-RL MAS, which transmits no confidence at all. An uncalibrated signal can mislead the downstream agent more than the absence of one. The confidence reward plays a complementary role: its removal leads to moderate accuracy drops but only slight Brier increases, indicating that it mainly improves the quality of the generated reliability analysis while the probability value is shaped more by the calibration loss. Replacing the readout with verbalized confidence degrades calibration across all scenarios, consistent with the RQ2 finding that discrete token sampling hinders stable and reliable probability calibration.

Confidence-aware weighting primarily affects search scenarios, where its removal reduces accuracy by 3--4pp; review scenarios are less affected, consistent with the scenario-level differences observed in the intervention analysis. All ablations have the smallest effect on math, where the clear structure of mathematical solutions allows the reviewing agent to judge correctness largely from the solution process. Each component's removal causes measurable degradation in at least one scenario or metric, confirming that calibration loss, confidence gradient, weighting mechanism, and readout design each serve a distinct and necessary function in MASCAL.

\FloatBarrier
\subsection{Generalization (RQ5)}
\label{sec:generalization}

We now examine whether MASCAL generalizes beyond the architecture and model scale used in the preceding experiments.

\textbf{Deeper collaboration chains.}
The sequential handoff in our main experiments involves a single upstream-downstream handoff. To test whether the learned confidence mechanism transfers to deeper architectures, we deploy the same checkpoints to 3-hop and 4-hop MuSiQue questions, forming deeper multi-hop retrieval-and-reasoning chains with two and three successive handoffs, each carrying a calibrated upstream confidence signal.

\begin{table}[!ht]
\centering
\caption{Accuracy (\%) and calibration on 3-hop and 4-hop MuSiQue questions, using checkpoints trained only on 2-hop data. Best results are \textbf{bolded}.}
\label{tab:deep_chain}
\setlength{\tabcolsep}{4pt}
\renewcommand{\arraystretch}{1.15}
\small
\resizebox{\columnwidth}{!}{%
\begin{tabular}{l ccc ccc}
\toprule
& \multicolumn{3}{c}{\textbf{3-hop}} & \multicolumn{3}{c}{\textbf{4-hop}} \\
\cmidrule(lr){2-4} \cmidrule(lr){5-7}
Method & Acc & Brier$\downarrow$ & AUROC$\uparrow$ & Acc & Brier$\downarrow$ & AUROC$\uparrow$ \\
\midrule
Base MAS           & 15.61 & 0.3771 & 0.6767 & 6.10 & 0.3149 & 0.6279 \\
Task-RL MAS        & 32.39 & 0.5209 & 0.6653 & 17.21 & 0.6737 & 0.5791 \\
MASCAL-Unweighted  & 34.51 & 0.1914 & \textbf{0.7764} & 19.65 & 0.1759 & 0.7439 \\
\textbf{MASCAL}    & \textbf{37.85} & \textbf{0.1903} & 0.7701 & \textbf{20.19} & \textbf{0.1741} & \textbf{0.7555} \\
\bottomrule
\end{tabular}%
}
\end{table}

MASCAL achieves the highest accuracy under both architectures (Table~\ref{tab:deep_chain}). The learned confidence communication, where the upstream conveys a calibrated reliability judgment and the downstream adjusts its retrieval and reasoning accordingly, remains effective in deeper chains unseen during training: each downstream node can leverage the upstream confidence to inform its own decisions.
The calibration gap further highlights the value of confidence training. The Brier score of Task-RL MAS deteriorates sharply with handoff count, far exceeding Base MAS at 4-hop. Without a reliability signal between agents, the downstream has no basis to discount unreliable upstream answers; these errors accumulate along the chain, and task-only training leaves the resulting predictions overconfident at every hop. MASCAL's Brier remains stable near the single-handoff level, and AUROC stays in the 0.74--0.78 range. The calibration ability acquired through training is not overfit to the training architecture but serves as a transferable capability that continues to function in longer retrieval chains.

\textbf{Model scale.}
To verify that the confidence learning mechanism does not depend on a specific model capacity, we replicate MASCAL with the same training recipe on Qwen3-1.7B and 4B, and compare alongside the 8B results from the main experiments.

\begin{table}[!ht]
\centering
\caption{Accuracy (\%) and calibration on sequential handoff across model scales. Best results per scale are \textbf{bolded}.}
\label{tab:model_scale}
\setlength{\tabcolsep}{3pt}
\renewcommand{\arraystretch}{1.15}
\small
\resizebox{\columnwidth}{!}{%
\begin{tabular}{l ccc ccc ccc}
\toprule
& \multicolumn{3}{c}{\textbf{1.7B}} & \multicolumn{3}{c}{\textbf{4B}} & \multicolumn{3}{c}{\textbf{8B}} \\
\cmidrule(lr){2-4} \cmidrule(lr){5-7} \cmidrule(lr){8-10}
Method & Acc & Brier$\downarrow$ & AUROC$\uparrow$ & Acc & Brier$\downarrow$ & AUROC$\uparrow$ & Acc & Brier$\downarrow$ & AUROC$\uparrow$ \\
\midrule
Base MAS           & 19.75 & 0.5964 & 0.4985 & 17.60 & 0.5794 & 0.6245 & 41.52 & 0.4111 & 0.6528 \\
Task-RL MAS        & 40.01 & 0.5537 & 0.5036 & 41.08 & 0.4367 & 0.6892 & 47.49 & 0.4498 & 0.6272 \\
MASCAL-Unweighted  & 44.50 & 0.2721 & 0.7000 & 45.73 & 0.2225 & 0.7460 & 55.45 & 0.2212 & 0.7297 \\
\textbf{MASCAL}    & \textbf{46.82} & \textbf{0.2205} & \textbf{0.7115} & \textbf{51.24} & \textbf{0.2190} & \textbf{0.7481} & \textbf{59.00} & \textbf{0.1947} & \textbf{0.8183} \\
\bottomrule
\end{tabular}%
}
\end{table}

MASCAL outperforms all baselines in both accuracy and calibration quality at every scale (Table~\ref{tab:model_scale}), confirming that calibration training and confidence-aware weighting are effective regardless of model capacity. The accuracy gain from confidence training grows with model scale, and calibration quality improves alongside it, suggesting that larger models benefit from the calibration objective on both dimensions.
The calibration contrast is especially pronounced at smaller scales: both Base MAS and Task-RL MAS at 1.7B achieve near-chance AUROC, indicating that without calibration training, smaller models produce almost no discriminative signal regardless of task optimization. MASCAL achieves substantially better Brier and AUROC even at 1.7B, showing that the calibration objective equips models with reliable probability estimation independent of model scale. Confidence-aware collaboration thus remains practical and viable even at smaller model scales.

\section{Conclusion and Future Work}

When multiple agents collaborate to retrieve and reason across Web content, the reliability of each agent's intermediate conclusions directly shapes downstream decisions.
We proposed MASCAL, which trains each agent to express calibrated answer confidence alongside its task and introduces confidence-aware weighting to train the downstream agent to act on this signal.
Across three collaboration architectures, MASCAL improves task accuracy and calibration quality, and the learned confidence communication transfers to deeper Web-retrieval chains without additional training, while the same training recipe remains effective at smaller model scales.
Intervention analysis and ablation experiments together show that neither producing a confidence signal nor exposing the downstream agent to it is sufficient on its own: the signal must be made reliable through dedicated calibration training, and the downstream agent must also learn to incorporate it into its decisions.

As agentic collaboration on the Web scales to more sources and longer reasoning chains, the reliability challenges facing collaborative systems grow as well.
The current study uses a fixed role assignment with all agents initialized from the same base model, and calibration relies on binary correctness feedback. Extending to heterogeneous MAS composed of different model families, and to open-ended generation tasks where correctness is not binary, calls for more flexible calibration objectives and end-to-end training across roles.
Extending confidence learning to multi-round interaction and dynamic role assignment is a natural next step. In such settings, confidence signals could inform system-level decisions such as role selection and computational allocation.
We plan to pursue these directions toward agentic systems in which calibrated confidence serves the full collaboration lifecycle, from each role's reliability judgment to system-level resource decisions, providing a more principled and reliable foundation for increasingly complex multi-agent information processing on the Web.

\bibliographystyle{ACM-Reference-Format}
\bibliography{references}
\appendix
\section{Experimental Details}
\label{sec:experimental_details}

In this appendix, we describe the retrieval setup, the training and evaluation protocols, and the prompts used in our experiments.

\subsection{Retrieval Infrastructure}

We use the Wiki-18 English Wikipedia dump released with Search-R1~\cite{searchr1} as the retrieval corpus.
To keep facts needed during training and evaluation reachable, we additionally index the candidate paragraphs supplied by MuSiQue, HotpotQA, and 2Wiki.
Retrieval is performed by E5-base-v2 over a flat inner-product index across the combined corpus.
Each call returns the top three passages, each truncated to the first 200 words.

Each trajectory is allocated at most three retrieval calls, split across roles differently in each architecture.
In sequential handoff, the upstream agent uses one call to answer the sub-question, and the downstream agent uses up to two more calls for the rest of the question.
In report fusion, each of the two upstream agents uses one call on the shared sub-question, and the downstream agent uses up to two calls after reading both reports.
The sub-question itself is produced by a Qwen3-8B planner used zero-shot, whose outputs are frozen for both training and evaluation.

\subsection{Training and Evaluation}

All four training scenarios share the hyperparameters in Table~\ref{tab:hparams}, with the scenario-specific choices described below.
Following prior work~\cite{credo}, we add H and L as reserved tokens whose embeddings are initialized from the existing \texttt{high} and \texttt{low} rows in the base vocabulary, giving the readout a semantically meaningful starting point.
The training set is split into two equal, disjoint halves: the upstream agent is trained on one half with the joint objective, then frozen, and generates confidence-bearing reports on the other half; the downstream agent is subsequently trained on these reports.

\begin{table}[h]
\centering
\caption{Training hyperparameters shared across scenarios.}
\label{tab:hparams}
\footnotesize
\setlength{\tabcolsep}{5pt}
\renewcommand{\arraystretch}{1.08}
\begin{tabular*}{\columnwidth}{@{\extracolsep{\fill}}ll@{\hspace{10pt}}ll@{}}
\toprule
Setting & Value & Setting & Value \\
\midrule
Base model      & Qwen3-8B         & Rollouts $G$                 & 8 \\
Optimizer       & AdamW (ZeRO-2)   & PPO clipping $\epsilon$      & 0.2 \\
Learning rate   & $2\times10^{-6}$ & Calibration warmup           & 30 steps \\
Epochs          & 3                & Weighting strength $\lambda$ & 0.5 \\
Inputs per step & 128              & Train sampling               & $T{=}1$, top-$p{=}1$ \\
\bottomrule
\end{tabular*}
\end{table}

The calibration loss weight is $\alpha{=}0.03$ on MuSiQue and $\alpha{=}0.01$ on LogiQA2 and DeepScaleR.
For report fusion, we train two upstream agents with random seeds 43 and 44 so that they produce independent reports on the shared sub-question.

\paragraph{Evaluation.}
Retrieval scenarios sample with $T{=}0.7$, top-$p{=}0.8$, top-$k{=}20$ and average over three seeds.
Review scenarios sample with $T{=}0.7$, top-$p{=}1.0$, top-$k$ off, drawing multiple independent samples per question.
Task correctness is measured by sub-EM~\cite{searchr1} for retrieval, \texttt{math-verify} for math, and exact choice matching for logic; confidence quality is measured by the Brier score and AUROC against the per-answer correctness label.

\subsection{Prompt Templates}
\label{sec:prompt_templates}

The system prompts for all roles are listed below.

\newtcolorbox{promptbox}[1]{%
  enhanced, breakable, before skip=4pt, after skip=4pt,
  colback=gray!5, colframe=gray!35, boxrule=0.3pt, arc=1.2pt,
  left=5pt, right=5pt, top=3pt, bottom=3pt,
  fonttitle=\footnotesize\bfseries, coltitle=black,
  colbacktitle=gray!18, attach boxed title to top left={xshift=4pt, yshift=-2pt},
  boxed title style={colframe=gray!35, boxrule=0.3pt, arc=1pt,
                     left=4pt, right=4pt, top=1pt, bottom=1pt},
  title={#1}}

\paragraph{Sequential handoff and report fusion.}
Both retrieval architectures use the same pair of prompts: the upstream agent answers its sub-question from retrieved documents, and the downstream agent continues with its remaining budget to answer the full question. In report fusion, the downstream input simply contains two upstream reports instead of one.

\begin{promptbox}{Retrieval, upstream agent}
{\footnotesize\ttfamily\raggedright\noindent
You are the first agent in a two-agent research team. You answer one sub-question of a larger question using the retrieved documents only; a teammate will use your report to finish the larger question. Reason briefly, then give your answer within <answer> </answer> tags. The answer must be as short as possible: only the entity, date, number or short phrase asked for, with no explanation and no extra qualifiers. Then cite at most two supporting sentences copied verbatim from the documents, each as <evidence doc="i"> sentence </evidence> where i is the document number. After the answer, analyze the uncertainty of your answer within <analysis> </analysis> tags: point out what could be wrong or ambiguous (missing or conflicting evidence, other possible readings); do not change your answer; if you find no uncertainty, say so explicitly. Then, within <confidence> </confidence> tags, output exactly one token: <CONF\_HIGH> if your answer is more likely correct than not, otherwise <CONF\_LOW>. Final format: brief reasoning, <answer> sub-answer here </answer> <evidence doc="i"> copied sentence here </evidence> <analysis> uncertainty analysis here </analysis> <confidence><CONF\_HIGH></confidence>.
\par}
\end{promptbox}

\begin{promptbox}{Retrieval, downstream agent}
{\footnotesize\ttfamily\raggedright\noindent
You are the second agent in a two-agent research team. A teammate has already answered the first sub-question of the question and sent you a report. The report covers only that sub-question; your job is to resolve the rest and answer the full question. Reason step by step. If you lack knowledge, you can call a search engine by <search> query </search>; it returns the top documents between <information> and </information>. Prefer facts found in retrieved documents over your own memory. Each search costs budget; you will be told how many searches remain. When you are ready, give the final answer to the full question within <answer> </answer> tags; it must answer the full question, never just the teammate's sub-question. The answer must be as short as possible: only the entity, date, number or short phrase asked for, with no explanation and no extra qualifiers. After the answer, analyze the uncertainty of your answer within <analysis> </analysis> tags: point out what could be wrong or ambiguous (missing or conflicting evidence, other possible readings); do not change your answer; if you find no uncertainty, say so explicitly. Then, within <confidence> </confidence> tags, output exactly one token: <CONF\_HIGH> if your answer is more likely correct than not, otherwise <CONF\_LOW>. Final format: reasoning and searches, then <answer> final answer here </answer> <analysis> uncertainty analysis here </analysis> <confidence><CONF\_HIGH></confidence>.
\par}
\end{promptbox}

\paragraph{Answer review.}
In the answer review scenarios, the upstream agent first solves the task on its own, and the downstream agent then examines the proposed solution before producing its own answer. The two roles use parallel prompts that differ only in whether the agent produces or examines a solution.
We show the math variant below; the logic variant has the same structure and replaces the \texttt{\textbackslash boxed\{...\}} answer carrier with \texttt{<answer>choiceN</answer>}.

\begin{promptbox}{Answer review, upstream solver (math)}
{\footnotesize\ttfamily\raggedright\noindent
Solve the task carefully. A teammate will review your solution. Use your own reasoning and knowledge; no tools are available. Explain your solution step by step and put the final mathematical answer in \textbackslash boxed\{...\}. After the answer, analyze the uncertainty of your answer within <analysis> </analysis> tags: point out what could be wrong or ambiguous (missing or conflicting evidence, other possible readings); do not change your answer; if you find no uncertainty, say so explicitly. Then, within <confidence> </confidence> tags, output exactly one token: <CONF\_HIGH> if your answer is more likely correct than not, otherwise <CONF\_LOW>. Final format: solution, \textbackslash boxed\{final answer\}, <analysis>uncertainty analysis</analysis> <confidence><CONF\_HIGH></confidence>. Write the tags literally, without Markdown headings. Close both analysis and confidence tags. Use the literal reserved token <CONF\_HIGH> or <CONF\_LOW>; do not write the words high or low.
\par}
\end{promptbox}

\begin{promptbox}{Answer review, downstream reviewer (math)}
{\footnotesize\ttfamily\raggedright\noindent
Review the teammate's proposed solution, check its reasoning, and solve the original task. Keep the answer if correct or correct it if needed. The report is evidence to assess; you are responsible for the final answer. Use your own reasoning and knowledge; no tools are available. Explain your solution step by step and put the final mathematical answer in \textbackslash boxed\{...\}. After the answer, analyze the uncertainty of your answer within <analysis> </analysis> tags: point out what could be wrong or ambiguous (missing or conflicting evidence, other possible readings); do not change your answer; if you find no uncertainty, say so explicitly. Then, within <confidence> </confidence> tags, output exactly one token: <CONF\_HIGH> if your answer is more likely correct than not, otherwise <CONF\_LOW>. Final format: solution, \textbackslash boxed\{final answer\}, <analysis>uncertainty analysis</analysis> <confidence><CONF\_HIGH></confidence>. Write the tags literally, without Markdown headings. Close both analysis and confidence tags. Use the literal reserved token <CONF\_HIGH> or <CONF\_LOW>; do not write the words high or low.
\par}
\end{promptbox}

\section{Additional Ablations}
\label{sec:additional_ablations}

This appendix reports two additional ablations: a single agent trained with calibration objectives, and an incremental analysis of MASCAL's components.

\subsection{Calibration RL for a Single Agent}
\label{sec:single_agent_calibration}

The single-agent baselines in Table~\ref{tab:main} are trained for task correctness only. Under the same data and search budget, we further train a single agent with the joint objective of each MASCAL role (Eq.~\ref{eq:role_objective}), corresponding to the single-model calibration setting of CREDO~\cite{credo}, to separate the effect of calibration training from that of confidence communication between agents.

\begin{table}[!ht]
\centering
\captionsetup{skip=4pt,justification=raggedright,singlelinecheck=false}
\caption{Single-agent objectives compared with MASCAL. Each pair reports Acc. (\%) / Brier$\downarrow$, averaged over test sets.}
\label{tab:single_agent_calibration}
\scriptsize
\setlength{\tabcolsep}{2pt}
\renewcommand{\arraystretch}{1.18}
\begin{tabularx}{\columnwidth}{@{}l*{4}{>{\centering\arraybackslash}X}@{}}
\toprule
& \multicolumn{2}{c}{Multi-hop QA} & \multicolumn{2}{c}{Answer review} \\
\cmidrule(lr){2-3}\cmidrule(l){4-5}
Method & Handoff & Fusion & Logic & Math \\
\midrule
Base Single-Agent & \multicolumn{2}{c}{36.07 / 0.3689} & 61.96 / 0.3472 & 60.41 / 0.3726 \\
Task-RL Single-Agent & \multicolumn{2}{c}{50.47 / 0.4275} & 71.54 / 0.1952 & 72.77 / 0.2513 \\
Calib-RL Single-Agent & \multicolumn{2}{c}{52.97 / 0.2119} & 72.53 / 0.1516 & 73.11 / 0.0847 \\
\addlinespace[2pt]
\rowcolor{methodBlue}
\textbf{MASCAL} & \textbf{59.00} / \textbf{0.1947} & \textbf{61.45} / \textbf{0.1940} & \textbf{76.59} / \textbf{0.1489} & \textbf{78.69} / \textbf{0.0816} \\
\bottomrule
\end{tabularx}
\end{table}

Calibration training brings the single agent's Brier scores close to MASCAL's without reducing accuracy, but its accuracy gains over Task-RL Single-Agent are small and remain well below MASCAL in every scenario.
A single agent's confidence has no consumer within the task; it turns into accuracy only when a downstream agent relies on it to search further or revise an answer.
Calibration training alone is therefore insufficient: MASCAL's accuracy gains rely on a collaboration mechanism that passes calibrated confidence to a downstream agent that acts on it.

\subsection{Component Analysis of MASCAL}
\label{sec:component_ablation}

The training ablation in Table~\ref{tab:training_ablation} removes components from the full MASCAL, but each MASCAL variant there still passes confidence between agents, so it cannot tell how much of the gain comes from calibration training itself and how much from transmitting the confidence.
We therefore take the opposite direction: starting from Task-RL MAS, we add MASCAL's three components one at a time to trace how each contributes to accuracy and calibration (Table~\ref{tab:component_ablation}):
\begin{itemize}[leftmargin=*,nosep]
\item \textbf{Cal.}: confidence policy gradient and BCE loss.
\item \textbf{Trans.}: upstream confidence included in the report.
\item \textbf{Wt.}: confidence-aware weighting in downstream training.
\end{itemize}

\begin{table}[!ht]
\centering
\captionsetup{skip=4pt,justification=raggedright,singlelinecheck=false}
\caption{Component analysis of MASCAL. Each pair reports Acc. (\%) / Brier$\downarrow$, averaged over test sets.}
\label{tab:component_ablation}
\scriptsize
\setlength{\tabcolsep}{2pt}
\renewcommand{\arraystretch}{1.18}
\begin{tabularx}{\columnwidth}{@{}ccc*{4}{>{\centering\arraybackslash}X}@{}}
\toprule
\multicolumn{3}{c}{Components} & \multicolumn{2}{c}{Multi-hop QA} & \multicolumn{2}{c}{Answer review} \\
\cmidrule(r){1-3}\cmidrule(lr){4-5}\cmidrule(l){6-7}
Cal. & Trans. & Wt. & Handoff & Fusion & Logic & Math \\
\midrule
 & & & 47.49 / 0.4498 & 54.78 / 0.3965 & 73.85 / 0.2266 & 72.91 / 0.2608 \\
$\checkmark$ & & & 53.28 / 0.2919 & 56.23 / 0.3016 & 73.89 / 0.1650 & 76.01 / 0.0918 \\
$\checkmark$ & $\checkmark$ & & 55.45 / 0.2212 & 57.26 / 0.1970 & 73.92 / 0.1517 & 78.32 / 0.0865 \\
\midrule
\rowcolor{methodBlue}
$\checkmark$ & $\checkmark$ & $\checkmark$ & \textbf{59.00} / \textbf{0.1947} & \textbf{61.45} / \textbf{0.1940} & \textbf{76.59} / \textbf{0.1489} & \textbf{78.69} / \textbf{0.0816} \\
\bottomrule
\end{tabularx}
\end{table}

With calibration training alone, the Brier score drops in every scenario, and accuracy rises clearly in sequential handoff and on math but changes little in report fusion and on logic.
The downstream agent is still less calibrated than the calibrated single agent in Table~\ref{tab:single_agent_calibration}, especially in search-based collaboration: it builds its final answer on an upstream answer whose reliability it cannot assess, and this uncertainty carries into its own confidence.
Answer review is less affected because the downstream agent can check the full proposed solution directly.
Passing the upstream confidence in the report closes most of this calibration gap and raises accuracy further, especially in sequential handoff and on math.
In report fusion and on logic, most of the accuracy gain appears only once confidence-aware weighting trains the downstream agent to act on the signal, and weighting also adds a clear gain in sequential handoff.
Calibration training thus improves confidence quality throughout and already lifts accuracy in some scenarios, whereas the remaining gains depend on passing this confidence downstream and training the downstream agent to use it.

\section{Training Overhead}
\label{sec:training_overhead}

Confidence-aware weighting replaces the upstream confidence with an unavailability marker and scores each generated downstream trajectory once more, introducing additional computation during training.
We compare MASCAL with MASCAL-Unweighted across the four scenarios and measure the complete upstream and downstream training-iteration time with and without confidence-aware weighting.
We run both variants on eight 80-GB GPUs using the training configuration in Appendix~\ref{sec:experimental_details}.

\begin{table}[!ht]
\centering
\captionsetup{skip=4pt,justification=raggedright,singlelinecheck=false}
\caption{Training-iteration time with and without confidence-aware weighting.}
\label{tab:training_overhead}
\footnotesize
\setlength{\tabcolsep}{3pt}
\renewcommand{\arraystretch}{1.12}
\begin{tabularx}{\columnwidth}{@{}l*{3}{>{\centering\arraybackslash}X}@{}}
\toprule
& \multicolumn{2}{c}{Time (h)} & \\
\cmidrule(lr){2-3}
Scenario & w/o weighting & MASCAL & Extra time \\
\midrule
Sequential handoff & 8.13 & 9.28 & +14\% \\
Report fusion & 9.95 & 10.29 & +3\% \\
Answer review (logic) & 9.61 & 10.45 & +9\% \\
Answer review (math) & 28.69 & 31.16 & +9\% \\
\bottomrule
\end{tabularx}
\end{table}

Across the four scenarios, confidence-aware weighting increases the total training-iteration time by only 3--14\%, since the extra pass scores existing trajectories rather than generating new ones.
In return, the downstream agent learns to act on upstream confidence, improving accuracy and calibration in every scenario (Table~\ref{tab:training_ablation}), with the largest accuracy gains in search-based collaboration.
Since the scoring pass only computes training weights, the downstream agent at deployment simply reads the confidence from the upstream report and answers without an extra scoring step.

\end{document}